%% file: localpolicy.tex
\documentclass[sigconf]{acmart}

\input{preamble/preamble}
\input{preamble/preamble_acronyms}

\usepackage{overpic}
\usepackage{xcolor}
\usepackage{cancel}

\usepackage{tikz}
\usetikzlibrary{shapes,decorations,arrows,calc,arrows.meta,fit,positioning}
\tikzset{
    -Latex,auto,node distance =1 cm and 1 cm,semithick,
    state/.style ={ellipse, draw, minimum width = 0.7 cm},
    point/.style = {circle, draw, inner sep=0.04cm,fill,node contents={}},
    bidirected/.style={Latex-Latex,dashed},
    el/.style = {inner sep=2pt, align=left, sloped}
}

\usepackage{parskip} 

\usepackage{amsthm,mathtools}
\usepackage{abbrevs}
\usepackage{amsmath}

\usepackage{bm}
\usepackage{dsfont}

\usepackage{wrapfig}
\usepackage{natbib} 

\definecolor{bblue}{rgb}{0.2,0.2,0.7}

\def\bl#1{{{{#1}}}}

\let\underbrace\LaTeXunderbrace

\newcommand{\argmax}{\mathop{\mathrm{argmax}}}
\newcommand{\E}{\mathbb{E}}

\newcommand\defeq{\mathrel{\overset{\makebox[0pt]{\mbox{\normalfont\tiny\sffamily def}}}{\, = \,}}}

\newcommand{\Prb}[1]{\mathbb{P}\left(#1\right)}
\newcommand{\kld}[2]{D_{\mbox{\tiny KL}} \big( #1 \ \vert\,\!\vert \ #2 \big)}
\newcommand{\jsd}[2]{D_{\mbox{\tiny JS}} \big( #1 \ \vert\,\!\vert \ #2 \big)}

\newcommand{\lpic}{{\textsc{lpi}}$^{\text{\sc cb}~}$}
\newcommand{\lpir}{{\textsc{lpi}}$^{\text{\sc rl}~}$}

\AtBeginDocument{%
  }

\setcopyright{acmcopyright}
\copyrightyear{2018}
\acmYear{2018}
\acmDOI{XXXXXXX.XXXXXXX}

\acmConference[Conference acronym 'XX]{Make sure to enter the correct
  conference title from your rights confirmation emai}{June 03--05,
  2018}{Woodstock, NY}
\acmPrice{15.00}
\acmISBN{978-1-4503-XXXX-X/18/06}

\begin{document}

%%
%% The "title" command has an optional parameter,
%% allowing the author to define a "short title" to be used in page headers.
\title{Local Policy Improvement for Recommender Systems }

%%
%% The "author" command and its associated commands are used to define
%% the authors and their affiliations.
%% Of note is the shared affiliation of the first two authors, and the
%% "authornote" and "authornotemark" commands
%% used to denote shared contribution to the research.

\author{Dawen Liang}
\affiliation{%
  \institution{Netflix}
  \country{}
}
\email{dliang@netflix.com}

\author{Nikos Vlassis}
\affiliation{%
  \institution{Adobe Research}
  \country{}
}
\email{vlassis@adobe.com}

%%
%% The abstract is a short summary of the work to be presented in the
%% article.
\begin{abstract}
Recommender systems predict what items a user will interact with next, based on their past interactions. The problem is often approached through supervised learning, but recent advancements have shifted towards policy optimization of rewards (e.g., user engagement). One challenge with the latter is policy mismatch: we are only able to train a new policy given data collected from a previously-deployed policy. The conventional way to address this problem is through importance sampling correction, but this comes with practical limitations. We suggest an alternative approach of local policy improvement without off-policy correction. Our method computes and optimizes a lower bound of expected reward of the target policy, which is easy to estimate from data and does not involve density ratios (such as those appearing in importance sampling correction). This local policy improvement paradigm is ideal for recommender systems, as previous policies are typically of decent quality and policies are updated frequently. We provide empirical evidence and practical recipes for applying our technique in a sequential recommendation setting.
\end{abstract}

%
% The code below is generated by the tool at http://dl.acm.org/ccs.cfm.
% Please copy and paste the code instead of the example below.
%


%%
%% Keywords. The author(s) should pick words that accurately describe
%% the work being presented. Separate the keywords with commas.

%%
%% This command processes the author and affiliation and title
%% information and builds the first part of the formatted document.
\maketitle

\input{sec_intro}
\input{sec_background}
\input{sec_method}
%\input{sec_related}
\input{sec_experiments}

\input{sec_conclusion}

%%
%% The next two lines define the bibliography style to be used, and
%% the bibliography file.
\bibliographystyle{ACM-Reference-Format}
\bibliography{localpolicy}

%%
%% If your work has an appendix, this is the place to put it.
\appendix
\input{appendix}

\end{document}

%% file: preamble/preamble.tex
\usepackage{lineno}

\usepackage{ragged2e}

\newcounter{parcount}

\DeclareRobustCommand{\parhead}[1]{\textbf{#1}~}
\definecolor{shadecolor}{gray}{0.9}

\usepackage{graphicx}
\usepackage[labelfont=bf]{caption}
\usepackage[format=hang]{subcaption}

\usepackage{booktabs, array, multirow}

\usepackage[algoruled]{algorithm2e}
\usepackage{algorithmic}
\usepackage{listings}
\usepackage{fancyvrb}
\fvset{fontsize=\small}

\usepackage[nameinlink]{cleveref}

\usepackage
[acronym,smallcaps,nowarn,section,nogroupskip,nonumberlist]{glossaries}
\glsdisablehyper{}
\lstdefinestyle{mystyle}{
    commentstyle=\color{OliveGreen},
    numberstyle=\tiny\color{black!60},
    stringstyle=\color{BrickRed},
    basicstyle=\ttfamily\scriptsize,
    breakatwhitespace=false,
    breaklines=true,
    captionpos=b,
    keepspaces=true,
    numbers=none,
    numbersep=5pt,
    showspaces=false,
    showstringspaces=false,
    showtabs=false,
    tabsize=2
}
\crefname{equation}{eq.}{eqs.}
\Crefname{equation}{Eq.}{Eqs.}
\creflabelformat{equation}{#1#2#3}

\usepackage{tikz}
\usetikzlibrary{bayesnet}
\usepackage[raggedright]{sidecap}
\sidecaptionvpos{figure}{t}

%% file: preamble/preamble_acronyms.tex
\newacronym{ELBO}{elbo}{evidence lower bound}
\newacronym{VAE}{vae}{variational autoencoder}
\newacronym{WMF}{wmf}{weighted matrix factorization}
\newacronym{BPR}{bpr}{{B}ayesian personalized ranking}
\newacronym{CDAE}{cdae}{collaborative denoising autoencoder}
\newacronym{NCF}{ncf}{neural collaborative filtering}
\newacronym{DAE}{dae}{denoising autoencoder}
\newacronym{MLP}{mlp}{multilayer perceptron}

\newacronym{IPS}{ips}{inverse propensity scoring}
\newacronym{RL}{rl}{reinforcement learning}
\newacronym{MDP}{mdp}{{M}arkov decision process}
\newacronym{LPI}{lpi}{local policy improvement}
\newacronym{TD}{td}{temporal-difference}

%% file: sec_intro.tex
% !TEX root = localpolicy.tex 

\section{Introduction}

Recommender systems are a ubiquitous component of the web. In the recommendation problem, we observe
how a set of users interact with a set of items, and our goal is to present these, as well as new, users a set of previously unseen items that they will enjoy. Historically this problem has been framed primarily as a prediction task via (self-)supervised learning \citep{koren2009matrix,steck2010training,liang2018variational,hidasi2015session,kang2018self,tang2018personalized,yuan2019simple,sun2019bert4rec}. In recent years, more research has been focusing on viewing recommendation as a form of \emph{intervention} \citep{schnabel2016recommendations,wang2020causal,bonner2018causal}, which naturally leads to a policy optimization approach to recommender systems in a decision making setting \citep{chen2019top,jeunen2021pessimistic,chen2022off,zou2019reinforcement,zhao2018deep,zhao2018recommendations,xin2020self,xin2022supervised,xin2022rethinking}. This also opens up 
the possibility to draw from the vast literature in the bandit and \gls{RL} research. 

However, there is one challenge in pursuing a policy optimization approach: Modern
recommender systems are mostly trained \emph{off-policy}, meaning we only observe feedback on items that have been recommended by the previously-deployed policy (known as bandit feedback \citep{swaminathan2015counterfactual}).
Hence, any inference about the unobserved feedback would involve a \emph{counterfactual} question: what would have happened had we deployed a different policy? 
A conventional approach to address this policy mismatch is through importance sampling correction. However, this approach suffers from large variance in practice, especially with large action spaces, which is often the case in recommender systems if we treat each item as an action a policy can execute. 

In this paper, we take a different approach to the problem of policy mismatch by locally optimizing a new policy. We draw from results in a number of fields including causal inference, bandits, and \gls{RL}, and present a method that computes and optimizes a lower bound of the expected reward of the new policy.
In essence we can perform policy optimization in an \emph{off-policy} fashion without resorting to importance sampling correction, and hence avoid the issue of large estimation variance. To the best of our knowledge, such perspective has not been applied to recommender systems before. We also present practical recipes and algorithms that can be applied to large-scale datasets with millions of users/interaction sequences. 
We argue that this type of local policy improvement is particularly well-suited for recommender systems: In a commercial recommender system, it is reasonable to assume the previously-deployed policy is of decent quality, and moreover the production systems tend to be frequently re-trained offline based on newly collected user feedback. Hence, being able to continuously improve the policy with theoretical guarantees makes recommender systems a perfect candidate to such a local policy optimization approach.

Furthermore, as part of the lower-bounding approach, we introduce a measurement which calculates how far the newly-obtained policy deviates from the logging policy. Combined with the standard offline ranking-based metrics commonly used in recommender systems evaluation, it provides us with a holistic view on the trade-off between exploitation (good offline metrics) and exploration (large deviation from the logging policy) in an offline setting. It is also able to precisely pinpoint the problem with importance sampling correction, which tends to overshoot and deviate too far from the logging policy without an easy way to control, as we demonstrate on both a semi-synthetic experiment as well as real-world datasets. In the meantime, we also show that our proposed approach can move between the two extremes with a single hyperparameter, giving practitioners the flexibility to make an informed decision before deploying a new policy online. 

In recent years, there have been works which combine recommender systems and \gls{RL}, among which \citet{xin2020self,xin2022supervised} are the closest to ours. Indeed, the objective functions introduced in \citet{xin2020self,xin2022supervised} are very similar to the local policy improvement objective that we arrive at. However, these approaches do not necessarily have the policy improvement guarantee as ours. In addition, despite the similarity on the objective function, empirically we demonstrate that our method can achieve significantly better results on two large-scale public e-commerce datasets. We further demonstrate the circumstances under which our approach outperforms the others by looking at where the most gains come from. 

This work draws some parallels to offline reinforcement learning \citep{levine2020offline} where an essential idea is \emph{pessimism}, effectively meaning to stay close to the data-generating policy \citep{kumar2020conservative,rashidinejad2021bridging,fujimoto2019off,brandfonbrener2021offline,wu2019behavior}. 

%% file: sec_background.tex
% !TEX root = localpolicy.tex 

\section{Background and related work} \label{sec:background}

We review offline learning for decision making (also known as batch learning from bandit feedback \citep{swaminathan2015counterfactual}) and connect it with recommender systems, specifically sequential recommendation. Throughout the paper, we follow the notational convention that upper-case letters (e.g., $X$, $A$, $R$) represent random variables and lower-case letters (e.g., $X=x$, $A=a$, $R=r$) represent actual values the random variables take. We sometimes write $X=x$ (and other random variables) simply as $x$ for brevity. 
We will also follow the convention that {random, data-driven} estimates have the form $\hat \cdot$ (e.g., $\hat{J}$ would be an estimate of $J$).

\subsection{Offline learning for decision making} \label{sec:background_offline}

In this paper, we consider the problem of offline learning from bandit feedback 
for decision making. Unlike online learning from bandit feedback, offline learning does not require online interaction and is capable of reusing existing logged data. 

We start with the bandit setting 
\citep{swaminathan2015counterfactual,ma2019imitation,jeunen2021pessimistic} with an offline dataset which consists of $n$ independent and identically distributed (i.i.d.) triplets $(x_i, a_i, r_i) \in \mathcal{X} \times \mathcal{A} \times \mathbb{R}$, $i=1,\dots,n$, representing the observed context, action taken (with $\mathcal{A}$ being the action space), and resulting reward, respectively. 
We use $(x,a,r)$ to refer to a generic draw of one of the above triplets, and assume it is sampled from the following generative process: $x \sim d(\cdot)$, $a \sim \Prb{\cdot \mid X=x}$, $r \sim p(\cdot \mid X=x, A=a)$.
The distribution $\mu(a\mid x)\defeq\Prb{A=a\mid X=x}$ is called the \emph{logging policy}. 
%Probabilities $\mathbb P$, expectations $\E$, and (co)variances without subscripts indicating otherwise are understood to be with respect to this distribution. For any function $f(x,a,r)$ we will write
% $
% \E_n[f(X,A,R)]=\frac1n\sum_{i=1}^nf(X_i,A_i,R_i).
% $

% In OPE, we are given a fixed policy $\pi(a\mid x)$ and are interested in estimating its average reward~$\E_{\pi}[R]$ using data collected under a logging policy $\mu(a\mid x)$. Here $\E_\pi$ refers to expectation under the distribution $X\sim \Prb{X},A\sim\pi(\cdot\mid X),R\sim\Prb{\cdot\mid A,X}$.
% The quantity $\E_{\pi}[R]$ can be interpreted as a counterfactual given that $\Prb{\cdot\mid A=a,X}$ is equal to the distribution of the \emph{potential} reward of slate~$a$ given context $X$, for every $a$; or, in other words, that $\Prb{\cdot\mid A,X}$ represents a structural model. This assumption, known as unconfoundedness or ignorability, is only needed for giving a causal interpretation to $\E_{\pi}[R]$ but not for estimation.

% The classical IS estimator for the problem of estimating $\E_{\pi}[R]$ is $\E_n\big[\frac{\pi(A\mid X)}{\mu(A\mid X)}R\big]$, which is unbiased but can suffer unacceptably large variance when $A$ is high-dimensional and $\mu$ provides coverage of all slates, as is necessary for the identification of arbitrary policies. 

% ===============================

In offline learning, given data collected from the logging policy $\mu$, our goal is to come up with a new policy $\pi$ that is ``good''. To quantify how good any arbitrary policy is, we adopt the notational shorthand for conditional mean reward $r(x, a) \defeq  \mathbb{E}[R \mid x, a]$ and  define the expected reward (also referred to as \emph{value} in the bandit and \gls{RL} literature) of a policy $\pi$ as the following expectation:
\begin{align}
J(\pi) \defeq  \E_{x \sim d(\cdot)} \, 
        \E_{a \sim \pi(\cdot \mid x)} \, 
        [r(x, a)]\label{eq:policy_value}
\end{align}
Given the above, we can formally define the objective of offline learning for decision making as $\argmax_\pi J(\pi)$. Although conceptually straightforward, this objective is difficult to optimize because it is essentially asking a \emph{counterfactual} question: what would have happened had we deployed policy $\pi$ instead of the logging policy $\mu$? This policy mismatch (often referred to as ``off-policy'') is what differentiates the offline learning setting from the online setting (also known as ``on-policy'') where we are able to test out any candidate policy and directly estimate its expected reward.

% For \emph{any} policy  $\mu(a \mid x)$ that is absolutely continuous wrt $\pi$ we can equivalently write
% %
% \begin{align}
% V(\pi) =
%          \E_{x \sim p(\cdot)} \, \E_{a \sim \bl{\mu(\cdot \mid x)}} \, 
%          \frac{\pi(a \mid x)}{\mu(a \mid x)} \ r(x,a)  
%          \label{eq:V2}
% \end{align}
% %

% \paragraph{Throwing the key idea early on!}

% Under absolutely continuity of $\pi$ wrt $\mu$, we can always write
% %
% \begin{align}
% \pi(a \mid x) \propto \mu(a \mid x) \ f(a,x)
% \end{align}
% %
% for some factors $f(a,x) \ge 0$, and the above reads
% %
% \begin{align}
% V(\pi) =
%          \E_{x \sim p(\cdot)} \, 
%          \frac{1}{f_\mu(x)} \,
%          \E_{a \sim \bl{\mu(\cdot \mid x)}} \, 
%          f(a,x) \ r(x,a)  
% \end{align}
% %
% where 
% \[
% f_\mu(x) \defeq \E_{a \sim \bl{\mu(\cdot \mid x)}} f(a,x).
% \]
% When $\mu$ and $f$ are available in closed-form, the normalizing function $f_\mu(x)$ can be computed analytically.

\parhead{Inverse propensity scoring for off-policy correction.}
% The above back-door and abs continuity assumptions offer a powerful framework for \emph{counterfactual} reasoning. For example, given data collected under $\mu$ 
% \[
%  (x_i, a_i, r_i) \sim p(\cdot) \, \bl{\mu(\cdot|x_i)} \, p(\cdot|x_i,a_i)
% \]
% we can \emph{estimate} the value of a target policy $\pi$ (Off-Policy Evaluation, OPE). The simplest such estimator (which is known to be minimax-optimal in the contextual bandit setting, but not in the multiarmed bandit setting REF) is Importance Sampling (IS):
%
We can re-write \Cref{eq:policy_value} using importance sampling: 
\begin{align}
J(\pi) &= \E_{x \sim d(\cdot)} \, 
        \E_{a \sim \pi(\cdot \mid x)} \, 
        [r(x, a)] \nonumber\\
        &=\E_{x \sim d(\cdot)} \,
        \E_{a \sim \mu(\cdot \mid x)} \,
        \biggl[\frac{\pi(a \mid x)}{\mu(a \mid x)} r(x, a) \biggl] \label{eq:policy_value_ips}
\end{align}
which effectively re-weights the data generated from $\mu$ to pretend it is sampled from $\pi$. This suggests the following \gls{IPS} estimator for $J(\pi)$:
\begin{align}
	\hat J(\pi) &= {\tfrac{1}{n} \sum_{i=1}^n \nolimits} \, 
 \frac{\pi(a_i \mid x_i)}{\mu(a_i \mid x_i)} \ r_i \label{eq:value_ips_estimator}
\end{align}
This estimator is unbiased, meaning given infinite amount of data, we will be able to estimate $J(\pi)$ accurately. 
Due to its simplicity, \gls{IPS} has become the \emph{de facto} choice to handle policy mismatch in both learning (especially in industrial setting~
\citep{bottou2013counterfactual,chen2019top,chen2022off}) as well as evaluation~\citep{strehl2010learning,saito2021evaluating}. 
However, the \gls{IPS} estimator in \Cref{eq:value_ips_estimator} is known to suffer from large variance when $\mu$ and $\pi$ are far from each other (i.e., when the importance weights $\frac{\pi(a_i \mid x_i)}{\mu(a_i \mid x_i)}$ take large values). It also requires that $\mu(a_i \mid x_i) > 0$ whenever $\pi(a_i \mid x_i) > 0$, a condition that if violated can lead to biased estimators~\citep{sachdeva2020off}. There are approaches that try to address some of these limitations, such as doubly robust estimators~\citep{dudik2011doubly}, variance regularization~\citep{swaminathan2015counterfactual}, weight clipping~\citep{strehl2010learning}, and self-normalized \gls{IPS}~\citep{swaminathan2015self}. However, all of them still carry the importance weights which can become problematic. In this paper, we take an alternative approach to locally improve the policy without resorting to importance sampling correction. 

%
% or, in the case of a known relationship between $\pi$ and $\mu$,
% \begin{align}
% \hat V(\pi) = {\tfrac{1}{n} \sum_{i=1}^n \nolimits} \,            \frac{f(a_i,x_i)}{f_\mu(x_i)} \ r_i  
%  \end{align}
% %
% [TODO: Make this precise.] The variance of the IS estimator scales roughly as $O(\frac{f^2(a,x)}{f_\mu(x)})$, which can behave well when ....

\subsection{Sequential recommendation as offline decision making}\label{sec:background_seq}
So far we have mostly talked about offline learning for decision making in a general setting in \Cref{sec:background_offline}. Now we contextualize it and connect to recommender systems, in particular sequential recommendation (also referred to as next item prediction).

As standard in recommender systems literature, we assume a set of users $u \in \mathcal{U}$ interacting with a catalog of items $a \in \mathcal{A}$. For the sake of concreteness, we assume the item catalog is the same as the action space of a bandit instance, as presented in \Cref{sec:background_offline}. For each user $u$, we have observed an interaction sequence $x_u = \{a_1, a_2, \dots, a_{|x_u|}\}$ (we will drop the subscript $u$ when there is no ambiguity) which is an ordered list of items. The goal of sequential recommendation is to build a model $f(\cdot \mid x)$ which takes input the interaction sequence $x$ and predicts the next item a user will likely interact with. Given the sequential nature of the interaction, it is common in practice to use a sequential model (e.g., convolutional or recurrent neural networks, transformers, etc.) as $f(\cdot \mid x)$ \citep{hidasi2015session,kang2018self,tang2018personalized,yuan2019simple,sun2019bert4rec} and train such a model via maximum likelihood given a collection of user-item interactions generated from some previously-deployed recommender system. 

Similar to how we define the bandit feedback dataset in \Cref{sec:background_offline}, we can assume the previously-deployed recommender system executes the ``logging policy'' $\mu$ (we do not need the analytical form for $\mu$, just assume that it exists). Furthermore, user-item interactions naturally induce some form of rewards (both positive and negative). For instance, in e-commerce use cases, the positive rewards can be clicks, add to cart, purchases, etc. The negative rewards are generally more implicit \citep{hu2008collaborative,pan2008one,liang2016modeling}. 
We can also interpret the standard sequential recommendation setting as having a 0-1 binary reward. %Nevertheless, the specific definition of rewards is an important yet orthogonal question and is beyond the scope of this paper. 
From this perspective, we can re-interpret sequential recommendation as an offline decision making problem, where the goal becomes: Given an offline dataset of triplets $\{(x_i, a_i, r_i)\}$ generated by the previously-deployed recommendation policy $\mu$, design a new policy $\pi$ so that when given as input an interaction sequence $x$ from a user, the recommender system recommends items $a\in\mathcal{A}$ following the policy $\pi(\cdot\mid x)$ to maximize the expected reward $J(\pi)$. 
Note that we can employ the same architectural choice for $\pi(\cdot \mid x)$ as we would do for $f(\cdot \mid x)$ in a standard sequential recommendation problem, as discussed above. %\citep{hidasi2015session,kang2018self,tang2018personalized,yuan2019simple,sun2019bert4rec}.

This offline decision-making perspective to recommender systems has been explored in recent years \citep{chen2019top,zhao2018deep,zhao2018recommendations,xin2020self,xin2022supervised,xin2022rethinking,chen2022off,zou2019reinforcement}. \citet{chen2019top,chen2022off} employ one-step off-policy \gls{IPS} correction with weight clipping to address policy mismatch in Youtube recommendation, a simplified version of which serves as a baseline in our experiment.
\citet{zhao2018recommendations} investigate how to incorporate negative feedback into \gls{RL}-based recommender systems. \citet{zhao2018deep} and \citet{zou2019reinforcement} incorporate \gls{RL} into specific recommendation settings. \citet{xin2020self,xin2022supervised,xin2022rethinking} are among the closest to our approach in this paper. 
However, despite the similarity in learning objective functions (especially with \citet{xin2020self,xin2022supervised}), the approaches in \citet{xin2020self,xin2022supervised,xin2022rethinking} do not necessarily enjoy the same theoretical justification as ours. We describe these approaches in more details in our experiment. 

%% file: sec_method.tex
% !TEX root = localpolicy.tex 

\section{Local policy improvement}

We present the main approach here. The basic idea is to optimize a \emph{lower bound} of the expected reward function $J(\pi)$ that is locally tight. Crucially, this lower bound is a function that is easy to estimate from data, and does \emph{not} involve density ratios (such as those appearing, e.g., in the \gls{IPS} approach). To the best of our knowledge, the proposed methodology has not been studied before in the context of recommender systems. 

We first present the approach in a simplistic setting where in the offline bandit feedback data we assume the context $x$ is sampled i.i.d. from a population distribution $x\sim d(\cdot)$ (akin to offline contextual bandit) and then we extend the result to more realistic settings where the context follows a Markov model (as in a Markov decision process). The distinction between the two settings is on the assumptions about the data generating process and consequently the expected reward associated with it.  

\subsection{Offline contextual bandit setting} \label{sec:penalty}

We take a slight detour and present a \emph{different} lower bound \citep{ma2019imitation} on the \gls{IPS}-corrected expected reward $J(\pi)$ in \Cref{eq:policy_value_ips} which both serves as a baseline and helps with deriving our learning algorithm later.  

Assume $r(x,a) \geq 0$ for all $(x,a) \in \mathcal{X} \times \mathcal{A}$.
(This is without loss of generality, since we can always shift $r(x,a)$ by a constant, without affecting the landscape of $J(\pi)$.) 
Applying the bound $x \geq 1 + \log x$ (which is tight at $x=1$) to the importance ratio $x = \frac{\pi}{\mu}$ at  \Cref{eq:policy_value_ips}, we get \citep{ma2019imitation}:
\begin{align}
\label{eq:Ma}
J({\pi}) \ \geq  \ \underbrace{\E_{x \sim d(\cdot)} \, \E_{a \sim \mu(\cdot \mid x)} \, \big[r(x,a) \, \log\pi(a \mid x) \big]}_{\mathcal{L}_\mu(\pi)} \ + \ \mbox{const} \end{align}
where `const' denotes terms that do not involve $\pi$. The lower-bounding function $\mathcal{L}_\mu(\pi)$ can be viewed as a surrogate function to optimize (over $\pi$) at the logging policy $\mu$. In \Cref{app:vlb} we also show that the same lower-bounding function $\mathcal{L}_\mu(\pi)$ can be derived from a different perspective \citep{vlassis2009model}, by bounding $\log J(\pi)$.

Optimizing the lower-bounding function $\mathcal{L}_\mu(\pi)$ w.r.t. $\pi$ (while keeping everything else fixed) we attain its maxima at
\begin{align} \label{eq:multi_r}
\pi(a \mid x) \propto \mu(a \mid x) \, r(x,a), \text{ for any }x \in \mathcal{X} \, .  
\end{align}
In practice we often do not have an explicit form for the logging policy $\mu$, making directly applying \Cref{eq:multi_r} difficult. Alternatively, the form of $\mathcal{L}_\mu(\pi)$ in \Cref{eq:Ma}  suggests the following estimator using data collected under $\mu$
\begin{align} \label{eq:wce}
	\hat{\mathcal{L}}_\mu(\pi) = {\tfrac{1}{n} \sum_{i=1}^n \nolimits} r_i \, \log \pi(a_i \mid x_i),
\end{align}
and optimizing it over $\pi$ leads to a \emph{reward-weighted cross-entropy} objective function which is amenable to stochastic optimization. 
% Asymptotically, this guarantees policy \emph{improvement}: 
% \begin{align*}
% \argmax_\pi  \hat{\mathcal{L}}_\mu(\pi) \overset{p}{\to} \pi^*,
% \qquad J(\pi^*) \geq J(\mu) \, .
% \end{align*}

% \begin{figure}
% {\centering
% \begin{overpic}[scale=.35,percent]{figs/emboundsimple.pdf}
% \put(70,49){\colorbox{white}{\small $J(\pi) \quad$}}
% \put(70,22){\colorbox{white}{\small $\mathcal{L}_\mu(\pi)+\mbox{const}$}}
% %\put(14.5,12){\colorbox{white}{\small $L(q_{\mu_2},\pi)$}}
% \put(25.2,4.7){\colorbox{white}{\small $\quad \mu \hspace*{45pt} \pi^* = \argmax_\pi  \mathcal{L}_\mu(\pi)$}}
% %\put(2,65){\colorbox{white}{\small $q_i(x) \propto p_{\theta_i}(x) x$}}
% \end{overpic}
% \par}
% \caption{A simple illustration of the lower-bound approach.}
% \label{fig:lower_bound_illustration}
% \end{figure}

\parhead{A direct lower bound.} \Cref{eq:Ma} provides a lower bound to the \gls{IPS}-corrected $J(\pi)$ in \Cref{eq:policy_value_ips}, which is tight at $\mu=\pi$, but it does not tell us what happens when $\pi$ deviates from the logging policy $\mu$. 
Here we present an alternative approach that \emph{directly} lower-bounds $J(\pi)$ in \Cref{eq:policy_value} by introducing a Kullback-Leibler (KL) divergence penalty between $\pi$ and $\mu$, averaged over the context $x\sim d(\cdot)$:
\begin{align}\label{eq:lowerbound_direct}
J(\pi) - \beta \cdot \mathbb{E}_{x\sim d(\cdot)}\Big[\kld{\pi(\cdot \mid x)}{ \mu(\cdot \mid x)}\Big] \leq J(\pi)
\end{align}
where $\beta$ is the Lagrangian multiplier and $\kld{\pi(\cdot \mid x)}{ \mu(\cdot \mid x)} \defeq \mathbb{E}_{a\sim\pi(\cdot\mid x)}[\log\pi(a\mid x) - \log \mu(a\mid x)]$. This is by definition a direct lower bound of the expected reward $J(\pi)$ because the KL divergence is always non-negative and the lower bound is tight at $\mu=\pi$.
Referring to the definition of $J(\pi)$ in \Cref{eq:policy_value}, we can re-write the lower bound in \Cref{eq:lowerbound_direct} (scaled by $1/\beta$) as:
\begin{align*}
&\mathbb{E}_{x\sim d(\cdot)}\mathbb{E}_{a\sim \pi(\cdot\mid x)}\Big[- \log \pi(a \mid x) + \big(\log \mu(a \mid x) + r(x, a)/\beta \big)\Big]\\
 =&~\mathbb{E}_{x\sim d(\cdot)}\Big[-\kld{\pi(\cdot \mid x)}{\mu(\cdot \mid x) \exp(r(x, a) / \beta)}\Big]
\end{align*}
In other words, maximizing the lower bound in \Cref{eq:lowerbound_direct} is equivalent to minimizing the KL divergence $\kld{\pi(\cdot \mid x)}{\mu(\cdot \mid x) \exp(r(x, a) / \beta)}$ for all $x\sim d(\cdot)$. In an unconstrained setting, this leads to the following solution 
\[
\pi(a \mid x) \propto \mu(a \mid x) \, \exp (r(x, a) / \beta), \text{ for any } x\in \mathcal{X}.
\]
In analogy with the bounding approaches discussed earlier that gave rise to \Cref{eq:wce} from \Cref{eq:multi_r}, the above suggests the following {weighted cross-entropy} as the \gls{LPI} objective:
%\nv{---let's discuss this `in analogy` argument---}
\[
\hat{\mathcal{L}}_\textsc{lpi} (\pi) = \frac{1}{n}\sum_{i=1}^n \nolimits
\exp (r_i / \beta) \, \log \pi(a_i \mid x_i) \, .
\]
This objective, though technically feasible, faces a practical concern
when the reward $r_i$ takes large value and/or %if 
$\beta$ is small. To mitigate the issue, we note that for any function $g(x)$, the multiplicative solution for the optimal $\pi$ can be re-written as:
\begin{align*}
\pi(a \mid x) &\propto \mu(a \mid x) \, \exp (r(x, a) / \beta) \, \\
&\propto \mu(a \mid x) \, \exp \big( (r(x, a) - g(x) ) / \beta \big)
\end{align*}
Here $g(x)$ can be as simple as $\max_a r(x, a)$ which ensures the weight $\exp((r(x, a) - g(x)) /\beta)$ does not exceed 1. 
A particularly interesting choice of $g(x)$ is $\mathbb{E}_{a\sim \mu(\cdot \mid x)}[r(x, a)]$ which is the conditional expected reward under $\mu$ for context $x$. We present a derivation in \Cref{app:policy_improv} that this choice of $g(x)$ leads to an objective function that is the lower bound on $J(\pi) - J(\mu)$, the gap on the expected reward between $\pi$ and $\mu$, which if maximized would lead to an \emph{improved} policy $\pi$ over the logging policy $\mu$. Denote $A_\mu(x,a)\defeq r(x, a) - \mathbb{E}_{a\sim \mu(\cdot \mid x)}[r(x, a)]$ as the \emph{residuals} of the predicted reward under $\mu$, we have the updated \gls{LPI} objective,
\begin{align} \label{eq:opi}
\hat{\mathcal{L}}_\textsc{lpi} (\pi) = \ \frac{1}{n}\sum_{i=1}^n \nolimits
\exp (\hat{A}_\mu(x_i,a_i) / \beta) \, \log \pi(a_i \mid x_i) \, .
\end{align}
We make a few remarks: 
\begin{enumerate}
\item The \gls{LPI} objective is \emph{off-policy} by design yet does not involve density ratios as in the \gls{IPS} approach and easy to estimate from data. Moreover, the Lagrangian multiplier $\beta$ gives us the flexibility to control exactly how far we want to deviate from the logging policy $\mu$. 
Furthermore, the KL divergence penalty term $\mathbb{E}_{x\sim d(\cdot)}\Big[\kld{\pi(\cdot \mid x)}{ \mu(\cdot \mid x)}\Big]$ can be estimated and used as a diagnostic measure for us to assess the deviation from the logging policy. 
\item The resulting objective in \Cref{eq:opi} is known in the literature under different names (e.g., \citep{peters2010relative,wang2018exponentially,peng2019advantage}). However, to the best of our knowledge, our derivation around lower-bounding $J(\pi)$, which is closely connected to how variational inference is presented in \citet{blei2017variational}, is new.
\item As mentioned above, the choice of function $g(x)$ does not have a direct impact in the contextual bandit setting. However, as we show next, setting $g(x)$ to $\mathbb{E}_{a\sim \mu(\cdot \mid x)}[r(x, a)]$ is also a convenient choice that generalizes to the
\gls{RL} setting.
\end{enumerate}

We close this section by noting a related \gls{IPS}-based objective function that has been used in practice (e.g., \citet{chen2019top}) as the following importance-sampling-corrected cross-entropy:
\begin{align}
    \hat{\mathcal{L}}_\textsc{ips}(\pi) = \frac{1}{n} \sum_{i=1}^n \nolimits
 \frac{\pi(a_i \mid x_i)}{\mu(a_i \mid x_i)} \
 r_i \ \log \pi(a_i \mid x_i) \, .
 \label{eq:ips_r}
\end{align}
The key difference to the weighted cross-entropy objectives that we derived in this section is that in \Cref{eq:ips_r} the weights of the $\log \pi(\cdot)$ terms also depend on $\pi$. It would be of interest to investigate whether, in analogy to the approaches presented above, \Cref{eq:ips_r} also admits an interpretation as an empirical lower bound of some related objective function.

% As a thought experiment, if we define $r'(x, a) \defeq \frac{\pi(a \mid x)}{\mu(a \mid x)}r(x, a)$, \Cref{eq:ips_r} becomes the reward-weighted cross-entropy objective \Cref{eq:wce} with rescaled rewards $r'$. If we plug it into \Cref{eq:optimal_p_ma}, we have 
% \[
% \pi(a \mid x) \propto \mu(a \mid x) \ r'(x, a) = 
% \pi(a \mid x) \, r(x, a)
% %\cancel{\mu(a \mid x)} \cdot \Big(\frac{\pi(a \mid x)}{ \cancel{\mu(a \mid x)}} \  r(x, a)\Big)
% \]
% which shows us exactly what \gls{IPS} aims to achieve: to remove the dependency on the logging policy $\mu$. This is in stark contrast with both the lower-bounding approach $\hat{\mathcal{L}}_\mu(\pi)$ and the penalty approach $\hat{\mathcal{L}}_\textsc{lpi} (\pi)$ where the dependency on $\mu$ can be explicitly derived. Both directions have their pros and cons but it is still interesting to see how they can be connected through the local policy improvement perspective.   

\subsection{The Markov decision process setting} \label{sec:rl}

% \PP throughout the paper we always consider offline setting with no online exploration/interaction. contextual bandit means i.i.d. context while RL means markovian state transition. 

We now extend the above results from the previous section to \gls{RL} setting with a \gls{MDP}. Besides the context/state $x\in \mathcal{X}$ (to be consistent with the previous section, we use the word ``context'' and ``state'' interchangeably in the \gls{MDP} setting), action $a\in \mathcal{A}$, and (immediate) reward $r(x, a)$, we also assume a discount factor $\gamma \in (0,1)$, a starting distribution $d_0(x)$, and a state transition model $x_t \sim \Prb{\cdot \mid x_{t-1}, a_{t-1}}$. A major difference between here and the prvious section is how a context/state is generated.

In an \gls{RL} setting, we are often presented with trajectories sampled from a policy $\pi$ as $\{(x_0, a_0, r_0), (x_1, a_1, r_1), \dots\}$.
We follow the standard notation of action-value $Q_\pi(x, a)$, state-value $V_\pi(x)$, and advantage $A_\pi(x, a)$, defined as
$Q_\pi(x, a) = \sum_{k=0}^\infty \gamma^{k} \, \mathbb{E}_\pi[r_{t+k}\mid X_t=x, A_t=a]$, $V_\pi(x) = \mathbb{E}_{a\sim\pi(\cdot\mid x)}[Q_\pi(x,a)]$, and $A_\pi(x, a) = Q_\pi(x, a) - V_\pi(x)$.
Define $d_\pi(x) \propto \sum_{t=0}^\infty \gamma^t \, \Prb{X=x_t \mid \pi}$ as the $\gamma$-discounted stationary distribution of states $x$ under policy $\pi$ \citep{sutton2018reinforcement}, where $\Prb{X=x_t \mid \pi}$ is the probability of being in state $x$ after following $\pi$ for $t$ timesteps. 
The expected (discounted) cumulative reward of a policy $\pi$ can be written as
\begin{align*}
J({\pi}) & \defeq \sum_{t=0}^\infty \gamma^t \, \E_\pi [r_t] = 
        \E_{x \sim d_\pi(\cdot)} \, \E_{a \sim \pi(\cdot|x)} \, [r(x,a)].
\end{align*}

% It can be shown that, for {any} function $f(x)$, we have {\bf (Achiam et al., 2017)}
% \begin{align}
% V({\pi}) = \E_{x \sim p(\cdot)} [ f(x) ] + 
% \frac{1}{1-\gamma} \E_{x \sim d_\pi(\cdot)} \, \E_{a \sim \pi(\cdot|x)} \, [ r(x,a) + \gamma \E_{x' \sim p(\cdot|x,a)} f(x') - f(x) ]
% \end{align}
% %
Recall in \Cref{sec:penalty} we have shown that the \gls{LPI} objective in \Cref{eq:opi} corresponds to a lower bound of the difference on the expected reward between two policies; a similar result exists for \gls{RL} \citep{Kakade02approximately,achiam2017constrained}
%and if we choose $f(x) = V_{\bl{\mu}}(x)$, we get {\bf (Kakade and Langford, 2002)}
%
\begin{align*}
J(\pi) - J(\mu) = \ \E_{x \sim d_\pi(\cdot)} \, \E_{a \sim \pi(\cdot|x)} \, [ A_{\bl{\mu}}(x,a) ] \, ,
\end{align*}
which, if maximized, would again lead to an \emph{improved} policy. However, the expectation over $d_\pi$ is problematic since we can neither compute it nor sample from it. Many modern \gls{RL} algorithms \citep{wang2018exponentially,peng2019advantage} optimize instead
\begin{align*}
\max_\pi \ \E_{x \sim d_\mu(\cdot)} \, \E_{a \sim \pi(\cdot|x)} \, [A_{\bl{\mu}}(x,a) ]
\end{align*}
subject to $\pi \approx \mu$, 
since now we can sample from $d_\mu$. From here, we can equivalently derive the same lower-bounding objective as in \Cref{sec:penalty} by introducing $\kld{\pi}{\mu}$ as a penalty term. This result has been known as relative entropy policy search \citep{peters2010relative} in the optimal control / \gls{RL} literature. Similar ideas are also explored in \citet{schulman2015trust,schulman2017proximal} for on-policy \gls{RL}.

%\nv{(I don't follow...)} 
It may appear that we have overloaded the definition of the expected reward $J(\pi)$ and advantage $A_\pi$ in both \Cref{sec:penalty} and this section. It is in fact an intentional choice since we can generalize our definition here to the offline contextual bandit setting by assuming $\gamma=0$ and the state $x\sim d(\cdot)$ are sampled independent of the previous state and action. To distinguish, we use \lpic and \lpir to refer to the offline contextual bandit and \gls{MDP} settings, respectively. We emphasize that the distinction between \lpic and \lpir is on what assumptions we have about the data generating process and consequently the expected reward associated with it.  

\parhead{Estimating advantage. } An important ingredient of our local policy improvement approach involves estimating the advantage function of a given policy $\mu$ (for the bandit setting, we can estimate it via the direct method \citep{dudik2011doubly}). There are mainly two different approaches: Monte-Carlo methods and \gls{TD} methods \citep{sutton2018reinforcement}. Given 
that in a typical recommender system setting
we only observe an incomplete sequence of user interactions (right-censored from the point when we collect the data), we primarily adopt the \gls{TD} approach, which introduces a loss function $\hat{\mathcal{L}}_\textsc{td}$. We keep the notation general since we can plug in different \gls{TD} estimators here.

\subsection{Practical considerations and recipes}\label{sec:recipes}

\parhead{Practical implications for recommender systems. } 
The local policy improvement perspective has been presented in various contexts within the bandit/\gls{RL}/optimal control literature over the years~\citep{Kakade02approximately,peters2010relative,ma2019imitation,peng2019advantage,wang2018exponentially,brandfonbrener2021offline,vlassis2009model,schulman2015trust,schulman2017proximal}. 
The essence of it is that if we are operating with a close proximity to the logging policy $\mu$, we are able to improve the policy locally with theoretical guarantees. 

In previous work, the logging policy has mostly been used as a way to debias the learning and evaluation through \gls{IPS} \citep{chen2019top,gilotte2018offline,schnabel2016recommendations}. To the best of our knowledge, there has not been an application of the local policy improvement approach in the recommender systems literature. This is rather surprising; we would argue that the local policy improvement perspective fits particularly well for recommender systems. 

In a typical recommender system, we deploy a production/logging policy, which generates new user interaction data. These data is collected and used to train a new model, and this process repeats on a regular basis. In this case, it is reasonable to assume the logging policy is likely of reasonable quality. Therefore, it is especially relevant if we can always train a new policy that is an \emph{improved} version of the previously-deployed policy. 

% \PP IPS only works well with smaller importance weights, in which case we know mu=pi, so here is an alternative. 

\begin{figure}[!ht]
  \centering
    \begin{subfigure}[b]{0.45\textwidth}
      \centering
  \includegraphics[width=0.75\columnwidth]{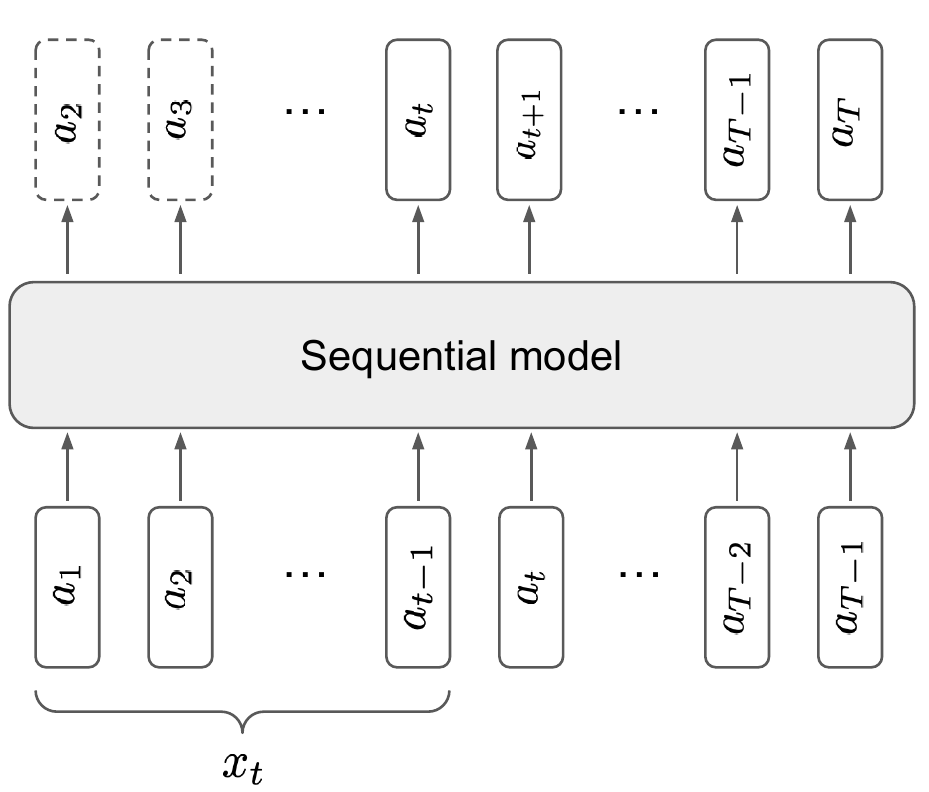}
        \caption{A general sequential recommendation architecture for policy optimization. The dashed boxes mean these actions can be skipped in the loss function, effectively allowing for longer input sequences $x$.}
        \label{fig:general_arch}
    \end{subfigure}
   \begin{subfigure}[b]{0.45\textwidth}
     \centering
  \includegraphics[width=0.9\columnwidth]{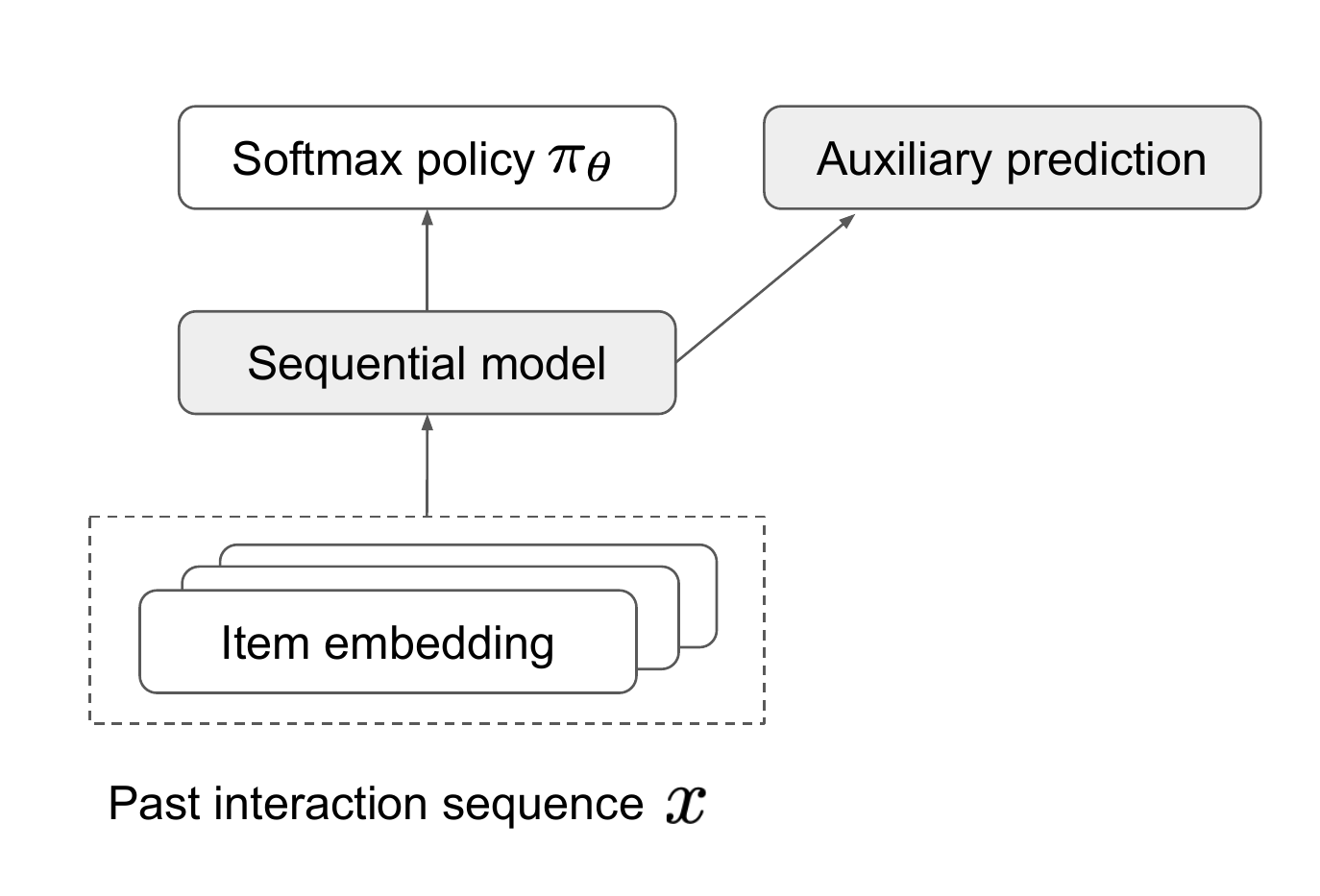}
        \caption{Policy parametrization with an auxiliary prediction head. }
        \label{fig:multi_head_arch}
    \end{subfigure}
    \caption{Diagrams of the general architecture and policy parametrization.}
    \label{fig:section3.3}
\end{figure}

\parhead{A general architecture. } \Cref{fig:general_arch} demonstrates how a typical sequential recommender system 
can be formulated from a policy optimization perspective presented in this paper. The backbone of the architecture is a sequential model, which can be a convolutional/recurrent neural network or a transformer \citep{hidasi2015session,kang2018self,tang2018personalized,yuan2019simple,sun2019bert4rec}. The input at each step is $x_t = \{a_1, \cdots, a_{t-1}\}$ and the output corresponds to action $a_t$. For certain architectures, if the input sequence is too long, making predictions and computing loss function across the entire sequence might be infeasible in one forward pass\footnote{Alternatively we can use a sliding window of fixed size and only make prediction at the last action. However, this way is much slower and in practice we found the metrics tend to be worse.}. Therefore, we allow the loss to be only computed over the most recent $m$ actions where $m < |x_T|$ (as demonstrated by the dashed boxes in \Cref{fig:general_arch}, meaning they can be skipped in the loss function). We use SASRec \citep{kang2018self}, one of the state-of-the-art baselines for sequential recommendation, as the backbone architecture for all the experiments. 

We have two separate objective functions to optimize: the local policy improvement part $\hat{\mathcal{L}}_\textsc{lpi}$ and the temporal-difference loss $\hat{\mathcal{L}}_\textsc{td}$. 
To parametrize the policy, we take a similar approach to some earlier work \citep{chen2019top,xin2020self,xin2022supervised} with a multi-head architecture shown in \Cref{fig:multi_head_arch}, which suggests the following general objective:
\begin{equation} 
\max_\theta \hat{\mathcal{L}}(\theta) = \hat{\mathcal{L}}_\textsc{lpi}(\theta) - \lambda \cdot \hat{\mathcal{L}}_\textsc{td}(\theta) \label{eq:general_obj}
\end{equation}
where $\lambda$ balances the two parts of the objective. The auxiliary prediction head is primarily optimized by \gls{TD} loss in our case, e.g., it can be a predicted action-value ${Q}_\theta(x, a)$ or state-value ${V}_\theta(x)$. One example for \gls{TD} loss is the Q-learning objective used in \citet{xin2020self,xin2022supervised} where the auxiliary head outputs one prediction for each action $a$ given state $x$ as ${Q}_\theta(x, a)$:
\begin{equation}
\hat{\mathcal{L}}_\textsc{td}(\theta) = \frac{1}{m} \sum_t\Big({Q}_\theta(x_t, a_t) - (r_t + \gamma \max_{a'} {Q}_\theta(x_{t+1}, a'))\Big)^2 \label{eq:q-learning}
\end{equation}
where the sum is over the last $m$ events as mentioned above. 
The auxiliary prediction head can also act as the weighting function (e.g., $\exp ({A}_\theta(x_i,a_i) / \beta)$) to the policy head, to which we cut the gradient and treat the weights as constant. This architecture also supports other related work \citep{chen2019top,chen2022off} by setting the auxiliary prediction head accordingly. The objective in \Cref{eq:general_obj} can be optimized with standard stochastic gradient and scale to large-scale datasets with millions of users/sequences. 

% \parhead{Data batching.} The objective in \Cref{eq:general_obj} is often optimized by mini-batch stochastic gradient. Given an input sequence $x$, we have two different options in terms of constructing the batch. The first is to break down $x$ into many sub-sequences, e.g., assume $x = [a_1, a_2, a_3]$, we can break it down into three (input sequence, action) pairs: $\bigl\{(\emptyset, a_1), ([a_1], a_2), ([a_1, a_2], a_3)\bigl\}$. For each sub-sequence, we only compute the loss at the last step and add them up. This is used in \citet{xin2020self,xin2022supervised}. An alternative option is to use the entire sequence to do one forward pass, collect loss after each step and add them up. Both options give unbiased stochastic gradient estimates. The difference is that given a fixed dataset, the first option would take more gradient steps per epoch while the second option would likely scale better to bigger datasets since it processes each sequence only once. In terms of empirical performance, it will likely be problem/dataset-dependent. In the experiment we use the second option.   

%% file: sec_experiments.tex
% !TEX root = localpolicy.tex 

\section{Experiments}
We evaluate the performance of the proposed local policy improvement approach to recommender systems both in a semi-synthetic offline contextual bandit setting and a \gls{MDP} setting with large-scale real-world e-commerce data. 

Evaluation for recommender systems is difficult because recommender systems operate under the partial feedback setup in which we only observe feedback on items that have been recommended by the deployed policy. One way to address this challenge is through  off-policy evaluation \citep{vlassis2019design,dudik2011doubly,precup2000eligibility,thomas2016data}. However, off-policy evaluation is still an active research area and it especially struggles with large action space, which is commonplace in recommender systems. 
%It is argued \citep{chen2022off} that offline (off-policy) evaluation  in general is more challenging than off-policy learning.  
%Online AB testing is often considered the gold standard in terms of evaluating a recommendation policy. However, even in AB testing, a recommendation policy is rarely deployed \emph{as is}; instead the results generated by a policy are often provided to a downstream model which might re-rank the items or construct a slate of items while considering other factors, e.g., business logic, and/or the diversity of the items in a slate. 

In this paper, we still primarily evaluate various policies following the standard procedure in sequential recommendation by looking at how next item is ranked by the new policy given the interaction history. We take this pragmatic approach since in practice before deploying a new recommendation policy online, traditional recommendation ranking metrics can still give us a rough guidance. Furthermore, as we demonstrate in this section, when we look at both the traditional ranking metrics as well as the deviation between the new policy and the logging policy, we are able to get a more holistic picture regarding the different approaches and make an explicit exploration (deviation from the logging policy)-exploitation (good offline ranking metrics) trade-off. 

In this section, we highlight the following results: First, in a semi-synthetic experiment under the offline contextual bandit setting, we demonstrate the challenge with traditional ranking metrics for evaluating recommender systems. Combined with the divergence penalty introduced in \Cref{sec:penalty}, we are able to pinpoint the problem with \gls{IPS}-based approach, which tends to overshoot and move too far from the logging policy with no easy way to control. We also demonstrate that our proposed \gls{LPI} approach is capable of shifting between staying very close to and moving very far away from the logging policy in a controllable manner, giving practitioners the option to make an explicit trade-off. 

Second, with two public e-commerce sequential recommendation datasets, we evaluate various approaches under the \gls{MDP} setting. We show that the proposed \gls{LPI} approach significantly outperforms other approaches with similar objective functions. Furthermore, we look into the circumstances where \gls{LPI} achieves better performance and identify that the gains are even more prominent with shorter sequences (i.e., contain less actions). 

\subsection{A semi-synthetic experiment} \label{sec:exp_cb}

We start with the relatively simpler offline contextual bandit setting in \Cref{sec:penalty} (i.e., we don't consider the transition of user-state as in a \gls{MDP}), since in this setting our local policy improvement approach \lpic is optimizing the lower bound on the gap $J(\pi) - J(\mu)$ without any approximation (cf. \gls{MDP}). 

\parhead{Dataset.} We conduct the experiments by re-purposing the Netflix Prize data, a user-movie rating dataset. The ratings are between 1 and 5. We only keep users who have
watched at least five movies. For each user, we sort all the ratings chronologically and take the most recent 200 ratings if there are more than 200 items in the interaction history. Given the sequence length of 200, we only compute the loss function over the most recent $m$ = 50 items but allow the sequence $x$ to be as long as 200, as discussed in \Cref{sec:recipes}. For rewards, we take ratings of 1 and 2 as $r=0$, ratings of 3 as $r=0.5$, and ratings of 4 and 5 as $r=1$. After preprocessing, we obtain a dataset with 53 million interactions from 472,987 user sequences and 17,770 items. We take 20,000 random user sequences as validation and test sets, respectively. 

\parhead{Experimental setup.} Given that we have defined both the positive ($r > 0$) and negative $(r = 0)$ rewards, for each policy $\pi$, we directly compute how much reward on average a greedy policy $\argmax_a \hat{\pi}(a \mid x)$ would acquire on heldout triplets $\{(x_i, a_i, r_i)\}_{i=1}^n$:
\[
AR@1 = \frac{1}{n} \sum_{i} r_i \cdot \mathds{1}[a_i = \argmax_a \hat{\pi}(a \mid x_i)]
\]
However, as mentioned earlier, with bandit feedback data we only observe partial feedback on items that have been recommended by the deployed policy. To simulate a setting where we have full reward information on every user-item pair, we create a semi-synthetic dataset by imputing the missing ratings using a matrix factorization model \citep{koren2009matrix}, which has proven to be successful at the original Netflix Prize Challenge.

It is known that the user rating data is missing-not-at-random \citep{marlin2007collaborative}, i.e., whether or not a user decides to rate an item depends on how much she likes this item. In essence, the more positive ratings are over-represented in the observational dataset collected organically from a recommender system. Therefore, a naive imputation approach will likely over-impute positive ratings than what would be expected from real-world data. To this end, we adopt the method developed by \citet{steck2010training} which introduces a baseline score for all the missing ratings and perform a weighted factorization over the entire user-item matrix. Once we have such an imputation model $\hat{r}_\textrm{MF}(x, a)$, we define an \emph{imputed} average reward (iAR) with respect to the same greedy policy as follows:
\[
iAR@1 = \frac{1}{n} \sum_{i} \hat{r}_\textrm{MF}(x_i, \argmax_a \hat{\pi}(a \mid x_i))
\]
In our local policy improvement approach, we introduce the divergence constraint $\kld{\pi(\cdot \mid x)}{\mu(\cdot \mid x)}$ into the objective function. Empirically, we can also use the \emph{data-driven} estimate $\kld{\hat{\pi}(\cdot \mid x)}{\hat{\mu}(\cdot \mid x)}$ as a measurement to reveal how far the learned new policy $\hat{\pi}(\cdot \mid x)$ and the estimated logging policy $\hat{\mu}(\cdot \mid x)$ are as a proxy to understand the degree of change in policy. However, KL divergence is asymmetric and hard to interpret. Instead, we report Jensen-Shannon (JS) divergence, which is symmetric and furthermore bounded between $0$ and $\log_e(2)\approx 0.69$:
\begin{equation}
\jsd{p}{q} = \frac{1}{2} \kld{p}{m} + \frac{1}{2} \kld{q}{m}
\end{equation}
where $m = (p + q)/2$. In our case, we can compute the Jensen-Shannon divergence between the learned new policy $\hat{\pi}(\cdot \mid x)$ and the estimated logging policy $\hat{\mu}(\cdot \mid x)$ for each context $x \sim d(\cdot)$ and compute an average over the dataset to estimate $\mathbb{E}_{x\sim d(\cdot)}[\jsd{\hat{\pi}(\cdot \mid x)}{\hat{\mu}(\cdot \mid x)}]$. We also compute $\mathbb{E}_{x\sim d(\cdot)}[\kld{\hat{\pi}(\cdot \mid x)}{\hat{\mu}(\cdot \mid x)}]$ and they show similar trend. To estimate the logging policy $\hat{\mu}$, we perform maximum likelihood estimation on the training interaction sequences without any reward. 

\parhead{Baselines.} We compare with the following baselines: 
\begin{itemize}
\item Logging($\hat{\mu}$): the estimated logging policy.
\item $\mathcal{L}_\mu(\pi)$: the policy that optimizes the reward-weighted cross-entropy $\mathcal{L}_\mu(\pi)$ which is shown to be the lower bound of the expected reward $J(\pi)$. 
\item \textsc{ips}: the importance-sampling corrected version of $\mathcal{L}_\mu(\pi)$ defined in \Cref{eq:ips_r}. We apply weight clipping with a threshold of $30$. Smaller threshold tends to have a reduced variance by introducing bias. 
\end{itemize}
% \textsc{ips} is similar to the approximation approach used in \citet{chen2019top} (without the top-$k$ correction) except here we are operating in a contextual bandit setting hence we only use the immediate reward $r$ as the weights as opposed to the discounted cumulative reward. 

All the policies follow the same architectural structure described in \Cref{sec:recipes}. For the backbone SASRec model, we use a 2-layer transformer with 8 attention heads over 200-dim embedding/hidden units. For \textsc{ips} we set the auxiliary prediction head to estimate the logging policy $\mu$ and cut the back-gradient as done in \citet{chen2019top} so that it will not interfere with the learning of the new policy. All the models are trained with Adam optimizer \citep{kingma2014adam} with batch size of 128 for 30 epochs. We pick the best hyerparameter/epoch by the heldout AR@1 on the validation set. 

\begin{table}[t]
\caption{Quantitative results on Netflix dataset. The standard errors on all three metrics are on the order of $10^{-4}$.}
  \centering
\begin{tabular}{ l c c c c }
   Policy & AR@1 & iAR@1 & $\jsd{\hat{\pi}}{\hat{\mu}}$  \\
  \toprule
  Logging ($\hat{\mu}$) & 0.060 & 0.696 & --- \\
  $\mathcal{L}_\mu (\pi)$ & 0.060 & 0.727 & 0.094 \\  
  \textsc{ips} & 0.030 & 0.690 & {\bf 0.504} \\
  \midrule
  \lpic ($\beta=0.001$) & 0.023 & 0.706 & 0.223 \\
  \lpic ($\beta=0.1$) &  0.061 & {\bf 0.739} & 0.118 \\
  \lpic ($\beta=1$) &  0.061 & 0.717 & 0.087 \\
  \lpic ($\beta=1000) $&  0.060 & 0.698 & 0.081 \\
  \bottomrule\\
\end{tabular}
\label{tab:netflix_reward}
\end{table}

\parhead{Experimental results and analysis.}
The results are summarized in \Cref{tab:netflix_reward}. First of all, the observational AR@1 are all very close except \textsc{ips} and \lpic with $\beta=0.001$. This is further supported by the much larger value of JS divergence for these two models than all the others, especially \textsc{ips} which clearly overshoots. This does not necessarily mean the \textsc{ips} or \lpic with $\beta=0.001$ will be terrible online. However, it does tell us that if we were to deploy them online, we would be faced with enormous amount of uncertainty. 

The imputed average reward tells a different story: all the approaches manage to achieve similar level of imputed average reward, including \textsc{ips} and \lpic with $\beta=0.001$. Given that the quality of the score depends heavily on the imputation model, we would not read too much into the exact ordering of various approaches. However, this still shows us the challenge with traditional ranking metrics for evaluating recommender systems: by observational offline metrics (AR@1), we would expect \textsc{ips} and \lpic with $\beta=0.001$ to perform poorly. Yet if we believe the imputed reward to be reasonably accurate, then all of these policies would have similar level of performance online. 

Looking at \lpic with different regularization strength $\beta$ on the KL divergence, we can see that first, as expected, with smaller $\beta$, the learned policy $\hat{\pi}$ starts to move away from the logging policy, but not so far away that both the observational AR@1 and the ``full-information'' iAR@1 are still maintained at a high level (unless with a very small $\beta=0.001$). Second, as $\beta$ gets larger, the learned policy starts to get close to the logging policy, reflected by the relatively closeness between $\hat{\mu}$ and \lpic (with $\beta=1000$) on iAR@1. 

These results demonstrate the versatility of our proposed local policy improvement approach where we can explicitly control the amount of policy increments depending on the specific use cases in practice.  

\subsection{The \gls{MDP} setting} \label{sec:exp_rl}

In the \gls{MDP} setting, we assume the user context/state $x$ follows a transition model $x_t \sim \Prb{\cdot \mid x_{t-1}, a_{t-1}}$. \gls{MDP} is a more realistic setting but also comes with more approximations as shown in \Cref{sec:rl}. Nevertheless, much of the experimental setup is still similar to \Cref{sec:exp_cb}.  

\parhead{Datasets}: We conduct experiments on two real-world e-commence datasets which have been used in recent work on applying \gls{RL} to recommender systems \citep{xin2020self,xin2022supervised,xin2022rethinking}\footnote{In \citet{xin2020self,xin2022supervised,xin2022rethinking} a subsample of 200,000 sequences is used for YooChoose dataset. In this paper we use the entire dataset with 4M sequences.}: \textbf{YooChoose} and \textbf{RetailRocket}. Both contain sequences of clicks and purchases/add-to-cart. For both datasets, we remove sequences with less than 3 clicks. For \textbf{RetailRocket} we further remove all the items that appear in less than 3 sequences.
The attributes of the datasets after preprocessing are in \Cref{tab:data}. 

\begin{table}
\begin{center}
\caption{Attributes of datasets for the \gls{MDP} setting after preprocessing. }
\begin{tabular}{ l c c c }
  \toprule
   & \textbf{YooChoose} & \textbf{RetailRocket}  \\
  \midrule
  \# of sequences & 4.43M & 195,523  \\
  \# of items & 48,255 & 70,852  \\
  \# of clicks & 24.6M & 1.18M\\
  \# of purchases & 1.15M & 57,269 \\
%   \midrule
%   \# of held-out users & 10,000 & 40,000 \\
  \bottomrule
\end{tabular}
\label{tab:data}
\end{center}
\end{table}

\parhead{Experimental setup.} We split all the sequences into training/validation/test sets with a 80/10/10 split. For each sequence we only keep the most recent 20 actions (>97\% of sequences in both datasets contain no more than 20 actions). We use the following two recommendation metrics: hit rate (HR) and normalized discounted cumulative gain (nDCG).
HR@$k$ is similar to recall@$k$, measuring whether the heldout item is in the top-$k$ of the recommended items. nDCG@$k$ also considers the rank of the heldout item, but it assigns
higher scores ${1}/{\log_2(\textrm{rank} + 1)}$ to higher positions.
Following \citet{xin2020self,xin2022supervised,xin2022rethinking}, we define both click-based and purchase-based metrics, and assign a higher reward ($r_\text{p}=1.0$) to actions leading to purchases compared to actions leading to only clicks ($r_\text{c}=0.2$).

In addition to HR@$k$ and nDCG@$k$, we again measure the JS divergence between the learned policy $\hat{\pi}$ and the estimated logging policy $\hat{\mu}$. Specifically, we can compute an average over the dataset to estimate $\mathbb{E}_{x\sim d_\mu(\cdot)}[\jsd{\hat{\pi}(\cdot \mid x)}{\hat{\mu}(\cdot \mid x)}]$.
%, which is the approximated divergence constraint that is added as a Lagrangian into the objective function.

\parhead{Baselines.} In additional to the existing baselines: the estimated logging policy $\hat{\mu}$ and the reward-weighted cross-entropy $\mathcal{L}_\mu(\pi)$, we also consider the following baselines:
\begin{itemize}
  \item Self-supervised Q-learning (\textsc{sqn}) \citep{xin2020self}: it optimizes an unweighted cross-entropy function  $\hat{\mathcal{L}}_\textsc{ce}(\pi)=\frac{1}{n}\sum_{i=1}^n\log\pi(a_i \mid x_i)$ regularized by \gls{TD} loss:
  $\hat{\mathcal{L}}_\textsc{sqn} = \hat{\mathcal{L}}_\textsc{ce} - \lambda \cdot \hat{\mathcal{L}}_\textsc{td}$.
  \item Self-supervised Actor-Critic (\textsc{sac}) \citep{xin2020self}: similar to \textsc{sqn}, this method optimizes an action-value-weighted cross-entropy function $\hat{\mathcal{L}}_\textsc{qce}(\pi) = \frac{1}{n}\sum_{i=1}^n \hat{Q}(x_i, a_i) \log \pi(a_i \mid x_i)$ (akin to actor-critic) regularized by the \gls{TD} loss:
   $\hat{\mathcal{L}}_\textsc{sac} = \hat{\mathcal{L}}_\textsc{qce} - \lambda \cdot \hat{\mathcal{L}}_\textsc{td}$.
  \item Policy Gradient (\textsc{pg}): similar to reward-weighted cross-entropy, but instead of immediate reward we use cumulative reward (referred to as ``reward-to-go''\citep{chen2021decision,janner2021offline}) as weight at each timestep. \textsc{pg} is by design an on-policy method. Here we sipmly treat it as a policy which optimizes a weighted cross-entropy objective. 
  \item Off-policy Policy Gradient (\textsc{ips-pg}): similar to \textsc{pg}, but we apply an one-step important-sampling correction $\frac{\pi(a_i \mid x_i)}{\mu(a_i \mid x_i)}$ \citep{chen2019top} which is shown to be biased but with much less variance. We further apply weight clipping with a threshold of 30. 
\end{itemize}

There has been some more recent approaches. Specifically, \citet{xin2022rethinking} considers a prompt-based approach (similar to \citet{chen2021decision} and \citet{janner2021offline} for offline \gls{RL}) to sequential recommendation. It is not obvious how we can parametrize such a policy so that we can compute the divergence between the learned policy and the logging policy, which is one of the main focuses in this paper. In addition, advantage function is also discussed in \citet{xin2022supervised}. However, the way it is computed slightly differs from the definition of the advantage function in the \gls{RL} literature and the added computation overhead is quite substantial, making it difficult to scale to the full \textbf{YooChoose} dataset. When we plug the advantage function defined in \citet{xin2022supervised} into our \gls{LPI} objective, we are able to achieve better results than what are reported in \citet{xin2022supervised}. However, we feel this is orthogonal to the contributions of this paper. We leave further investigation to future work.

We implement all the baselines following the same architecture and parametrization as in \Cref{fig:multi_head_arch} to eliminate any difference in implementation details. As for the architecture of the backbone SASRec model, we adopt the same hyperparameters from \citet{xin2020self,xin2022supervised} with a single-layer transformer with one attention head over 64-dim embedding/hidden units. For the \gls{TD} loss, we adopt the Q-learning objective in \Cref{eq:q-learning} to optimize the auxiliary prediction head. To stabilize the Q-learning, we adopt the double Q-learning technique \citep{hasselt2010double} which proves very helpful. For \lpir, we compute the advantage analytically from the estimated action-value $\hat{Q}$.  All the models are trained with Adam optimizer with batch size of 256 for 20 epochs. We pick the best hyerparameter/epoch by $r_\text{p} \cdot \text{nDCG}_\text{p}@20 + r_\text{c} \cdot \text{nDCG}_\text{c}@20$ on the validation set where nDCG$_\text{p}$ and nDCG$_\text{c}$ are purchase and click nDCG, respectively. 

\begin{table*}[t]
  \caption{Quantitative results on YooChoose and RetailRocket. The largest numbers are boldfaced. The standard errors on hit rate and nDCG are listed in the table. The asterisk ($^*$) on hit rate and nDCG indicates the difference between the largest and second largest numbers are statistically significant at $\alpha=0.05$ level with a paired t-test. }
  \label{tab:results_rl}
  %\resizebox{\textwidth}{!}{%
  \begin{subtable}[h]{\textwidth}
    \begin{tabular}{c cccccc cccccc c}
      %\toprule
       &  \multicolumn{6}{c}{ Purchase ($\pm 0.002$) }  &  \multicolumn{6}{c}{ Click ($\pm <0.001$) } & \\
         \cmidrule(lr){2-7} \cmidrule(lr){8-13}
          &  \multicolumn{3}{c}{Hit Rate} & \multicolumn{3}{c}{nDCG} & \multicolumn{3}{c}{Hit Rate} & \multicolumn{3}{c}{nDCG} & $\jsd{\hat{\pi}}{\hat{\mu}}$ \\
         Policy & @5 & @10 & @20 & @5 & @10 & @20 & @5 & @10 & @20 & @5 & @10 & @20 & $\leq \log_{\mathrm{e}}(2)$\\
        \cmidrule(lr){1-1}     \cmidrule(lr){2-4} \cmidrule(lr){5-7} \cmidrule(lr){8-10} \cmidrule(lr){11-13} \cmidrule(lr){14-14} 
         $\hat{\mu}$ &  0.317 & 0.411 & 0.493 & 0.227  & 0.258  & 0.279 & 0.464 & 0.552 & 0.628 & 0.376 & 0.404 & 0.423 & ---\\
        $\mathcal{L}_\mu(\pi)$ &  0.322 & 0.415 & 0.496 & 0.232 & 0.262 & 0.283 & 0.461 & 0.547 & 0.623 & 0.373 & 0.401 & 0.420 & 0.089 \\
        \textsc{pg} & 0.318 & 0.411 & 0.490 & 0.227 & 0.257 & 0.277 & 0.457 & 0.543 & 0.617 & 0.371 & 0.399 & 0.417 & 0.095  \\
        \textsc{ips-pg} & 0.245 & 0.295 & 0.344 & 0.187 & 0.203 & 0.215 & 0.100 & 0.123 & 0.148 & 0.076 & 0.083 & 0.089 & {\bf 0.526}  \\
        \midrule
        \textsc{sqn} & 0.366 & 0.460 & 0.539 & 0.266  & 0.296  & 0.316 & 0.479 & 0.567 & 0.642 & 0.387  & 0.415  & 0.434 & 0.159 \\
        \textsc{sac} & 0.364 & 0.458 & 0.537 & 0.266  & 0.297  & 0.317 & 0.473 & 0.561 & 0.637 & 0.384  & 0.413  & 0.432 & 0.167 \\
        \lpir & {\bf 0.372}$^*$ & {\bf 0.469}$^*$ & {\bf 0.547}$^*$ & {\bf 0.271}$^*$  & {\bf 0.303}$^*$  & {\bf 0.323}$^*$ & {\bf 0.479}$^*$ & {\bf 0.567}$^*$ & {\bf 0.643}$^*$ & {\bf 0.388}$^*$  & {\bf 0.417}$^*$  & {\bf 0.436}$^*$ & 0.156\\
      % \bottomrule
    \end{tabular}
    \newline
    \caption{YooChoose}
  \label{tab:results_rc}
  \end{subtable}

  \begin{subtable}[h]{\textwidth}
    \begin{tabular}{c cccccc cccccc c}
      %\toprule
       &  \multicolumn{6}{c}{ Purchase ($\pm 0.007$) }  &  \multicolumn{6}{c}{ Click ($\pm0.002$) } & \\
         \cmidrule(lr){2-7} \cmidrule(lr){8-13}
          &  \multicolumn{3}{c}{Hit Rate} & \multicolumn{3}{c}{nDCG} & \multicolumn{3}{c}{Hit Rate} & \multicolumn{3}{c}{nDCG} & $\jsd{\hat{\pi}}{\hat{\mu}}$ \\
         Policy & @5 & @10 & @20 & @5 & @10 & @20 & @5 & @10 & @20 & @5 & @10 & @20 & $\leq \log_{\mathrm{e}}(2)$\\
        \cmidrule(lr){1-1}     \cmidrule(lr){2-4} \cmidrule(lr){5-7} \cmidrule(lr){8-10} \cmidrule(lr){11-13} \cmidrule(lr){14-14} 
        $\hat{\mu}$ &  0.631 & 0.669 & 0.698 & 0.551  & 0.563  & 0.571 & 0.445 & 0.494 & 0.537 & 0.374  & 0.390  & 0.401 & ---\\
        $\mathcal{L}_\mu(\pi)$ & 0.631 & 0.663 & 0.689 & 0.561  & 0.572  & 0.578 & 0.440 & 0.487 & 0.528 & 0.373  & 0.388  & 0.398 & 0.282\\
        \textsc{pg} & 0.621& 0.660 & 0.687 & 0.545  & 0.558  & 0.565 & 0.444 & 0.491 & 0.533 & 0.375  & 0.391  & 0.401 & 0.282 \\
        \textsc{ips-pg} & 0.257 & 0.270 & 0.285 & 0.224 & 0.228 & 0.232 & 0.107 & 0.117 & 0.127 & 0.089 & 0.093 & 0.095 & {\bf 0.571}\\
        \midrule
        \textsc{sqn} & 0.663 & {\bf 0.702} & 0.723 & 0.593 & 0.605  & 0.611 & 0.468 & 0.520 & {\bf 0.567} & 0.396  & 0.413  & 0.424 & 0.270 \\
        \textsc{sac} & 0.654 & 0.689 & 0.716 & 0.578  & 0.589  & 0.596 & 0.462 & 0.512 & 0.556 & 0.393  & 0.409  & 0.421 & 0.301 \\
        \lpir & {\bf 0.665} & 0.700 & {\bf 0.724} & {\bf 0.599}$^*$   & {\bf 0.610}$^*$  & {\bf 0.616}$^*$  & {\bf 0.471}$^*$  & {\bf 0.522} & {\bf 0.567} & {\bf 0.400}$^*$   & {\bf 0.416}$^*$   & {\bf 0.428}$^*$  & 0.273\\
    %   \bottomrule
    \end{tabular}
    \newline
    \caption{RetailRocket}
    \label{tab:results_rr}
    \end{subtable}
    %}
\end{table*}

\parhead{Experimental results and analysis.}
The results are summarized in \Cref{tab:results_rl}. First of all, we can see that \textsc{sqn}, \textsc{sac}, and \lpir all outperform the logging policy and $\mathcal{L}_\mu(\pi)$. 
Secondly, \lpir significantly outperform \textsc{sqn} and \textsc{sac}, especially on the large-scale \textbf{YooChoose} dataset. On \textbf{RetailRocket}, the standard errors on the metrics are bigger due to the smaller sample size. However, we can still see that, for \lpir, all the nDCGs, which assign larger weights to higher positions, are consistently better. Considering the similarity in the objective functions between \lpir and \textsc{sqn}/\textsc{sac}, the difference in performance is striking.  
It is also interesting to note that reward-weighted cross-entropy $\mathcal{L}_\mu(\pi)$, which does not directly control how far to move away from the logging policy, can have a sizable swing in terms of the deviation across datasets (much smaller JS divergence than \lpir on \textbf{YooChoose} while larger on \textbf{RetailRocket}). 

Finally, similar to \Cref{sec:exp_cb}, we see that \textsc{ips-pg} achieves lower metrics and moves far away from the logging policy with the highest JS divergence. Again, this does not necessarily mean the policy learned with \textsc{ips-pg} would be bad online, but the uncertainty is much higher compared with all the other policies considered here.  

\begin{figure}[!ht]
  \centering
    \begin{subfigure}[b]{\columnwidth}
      \centering
  \includegraphics[width=\columnwidth]{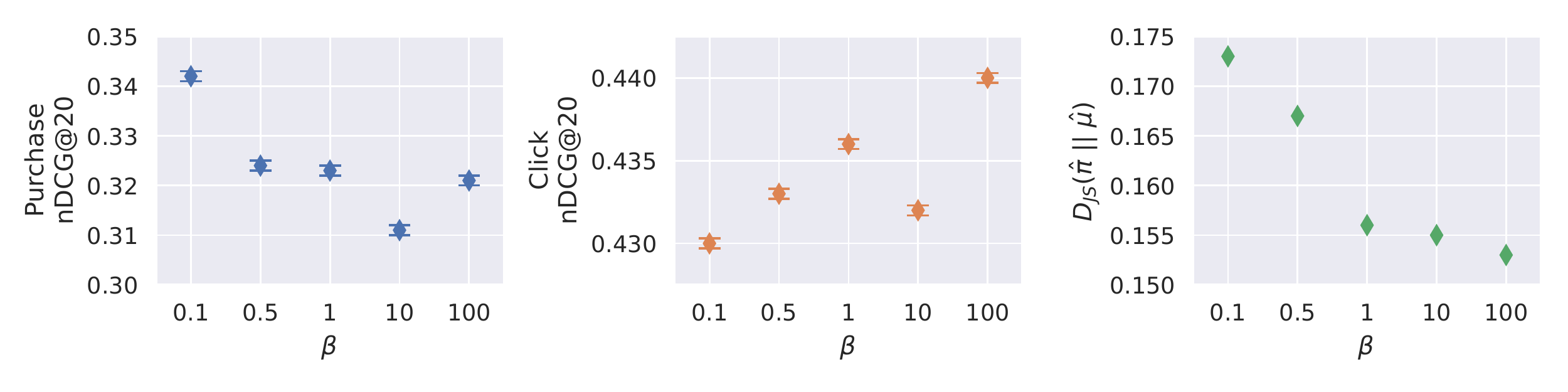}
        \caption{Varying $\beta$'s.}
        \label{fig:beta}
    \end{subfigure}
   \begin{subfigure}[b]{\columnwidth}
     \centering
  \includegraphics[width=\columnwidth]{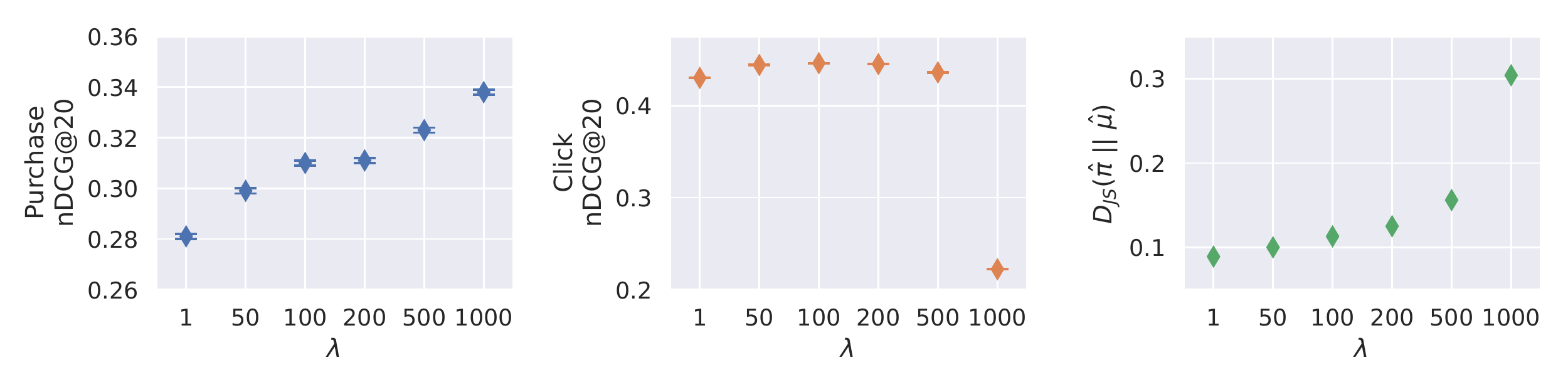}
        \caption{Varying $\lambda$'s.}
        \label{fig:lam}
    \end{subfigure}
    \caption{The YooChoose validation nDCG@20 with varying hyperparameters.}
    \label{fig:rl_exp}
\end{figure}

\parhead{The effect of hyperparameters. } We examine the two hyperparameters: the Lagrangian multiplier on the KL divergence $\beta$ and the regularization strength on the \gls{TD} loss $\lambda$. The validation metrics on \textbf{YooChoose} are shown in \Cref{fig:rl_exp}. For $\beta$'s, we can see that with larger value, the JS divergence between the learned policy $\hat{\pi}$ and the estimated logging policy $\hat{\mu}$ goes down, as expected. Interestingly it also often corresponds to an increase in the click metrics. This is understandable given the logging policy is trained with equal weights on clicks and purchases and there are many more clicks in the dataset. For $\lambda$'s, we can see a negative correlation between purchase and click metrics with increasing $\lambda$'s. Furthermore, larger $\lambda$ has a similar effect as smaller $\beta$, meaning $\lambda$ also has the capability of moving the policy away from the logging policy. We find that $\lambda$'s also play a very important role with \textsc{sqn} and \textsc{sac} as well. 

\begin{figure}[!ht]
  \centering
    \begin{subfigure}[b]{\columnwidth}
      \centering
  \includegraphics[width=\columnwidth]{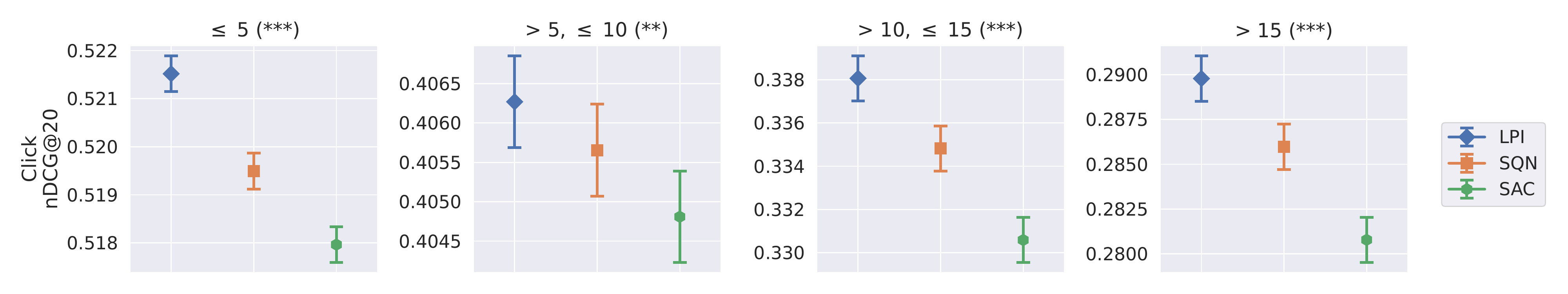}
        \caption{Sequence-level click nDCG by the number of clicks per sequence.}
        \label{fig:click_ndcg_yoochoose}
    \end{subfigure}
   \begin{subfigure}[b]{\columnwidth}
     \centering
  \includegraphics[width=\columnwidth]{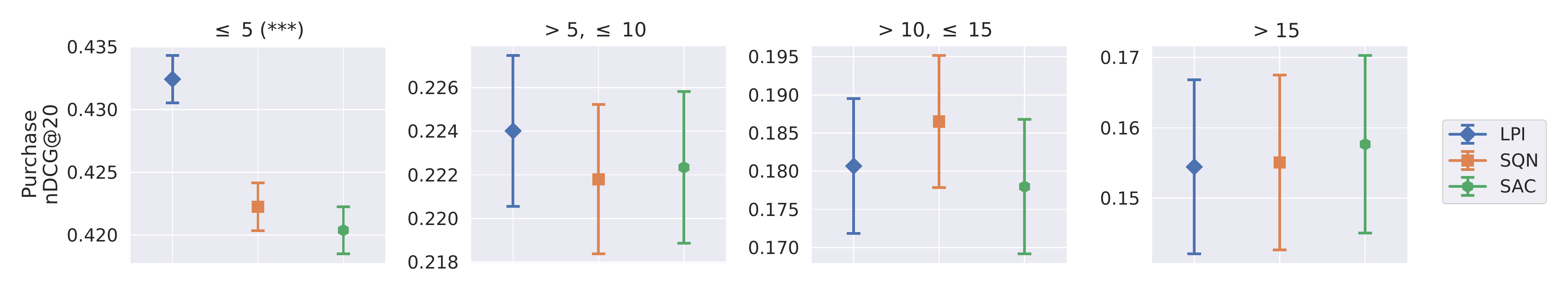}
        \caption{Sequence-lever purchase nDCG by the number of purchases per sequence.}
        \label{fig:purchase_ndcg_yoochoose}
    \end{subfigure} \
    \caption{Sequence-level validation metric breakdown by action counts on YooChoose. For each subplot, a paired t-test is performed and * indicates statistical significance at $\alpha=0.05$ level, ** at $\alpha=0.01$ level, and *** at $\alpha=0.001$ level. }
    \label{fig:ndcg_yoochoose_breakdown}
\end{figure}

\parhead{Where do the gains come from?} To further investigate under what circumstances \gls{LPI} achieves better performance, we aggregate sequence-level validation click/purchase metrics and break them down by the number of clicks/purchases per sequence. The results on \textbf{YooChoose} are illustrated in \Cref{fig:ndcg_yoochoose_breakdown}. (The results for \textbf{RetailRockets} are similar and can be found in \Cref{app:retailrocket_breakdown}.) Specifically, we count the number of clicks/purchases per sequence and break down sequence-level average click/purchase nDCG@20 into buckets. We can see that \gls{LPI} consistently outperforms other methods on the bucket with $\leq$5 clicks/purchases. Especially with purchase, all three methods are on par with each other on the rest of the three buckets. This shows that \gls{LPI} is more effective when there have not been many actions in the sequence yet. 

%% file: sec_conclusion.tex
% !TEX root = localpolicy.tex 

\section{Conclusion}

In this paper, we present a local policy improvement approach for recommender systems, bringing in results from various literature threads. We are able to optimize for a new policy in an off-policy fashion without resorting to \gls{IPS} (which is known to suffer from large variance). We argue that this perspective is particularly well-suited in recommender systems, and we experimentally demonstrate the flexibility of the proposed approach. We also show that empirically the proposed \gls{LPI} approach significantly outperforms other similar approaches. Furthermore, we observe that the gains from \gls{LPI} are especially prominent with shorter sequences (i.e., contain less actions). We hope that the presented methodology will motivate researchers to consider alternative ways to optimize recommender systems policies.

%% file: appendix.tex
% !TEX root = localpolicy.tex 
\newpage
\section{Appendix}
\subsection{A variational lower bound on $\log J(\pi)$} \label{app:vlb}
%%%%%%%
In this section we show that the lower-bounding function $\mathcal{L}_\mu(\pi)$ in \Cref{eq:Ma} can be derived from a different perspective \citep{vlassis2009model}, by bounding $\log J(\pi)$.

Assuming again $r(a,x) \ge 0$, for \emph{any} policy $\varpi(a \mid x)$ we define the following joint distribution on $(x,a) \in \mathcal{X} \times \mathcal{A}$
\begin{align}
q_\varpi(x,a) \, \defeq \, \frac{d(x) \, \varpi(a \mid x) \, r(a,x)}{J(\varpi)} \, .\nonumber
\end{align}
By setting $\varpi=\pi$, taking logs, and re-arranging terms, we get:
\begin{align}
\label{eq:logVqpi}
\log J(\pi) &= \log \pi(a \mid x) - \log q_\pi(x,a)
+ \mbox{const}, \ \ \forall (x,a).
\end{align}
Now, taking expectation of both sides of \Cref{eq:logVqpi}  w.r.t. {$q_\varpi$} and re-arranging terms, we end up with the following ``free energy'' function:
\begin{multline}   
\label{eq:Lvarpi}
\mathcal{F}(\pi, q_\varpi)=
\log J(\pi) - D_{\mbox{\tiny KL}} \big( q_\varpi \ \vert\,\!\vert \ q_\pi \big)  \\
= \frac{1}{J(\varpi)} \
\underbrace{\E_{x \sim d(\cdot)} \E_{a \sim \varpi(\cdot \mid x)} \big[r(x,a) \log \pi(a \mid x) \big]}_{\mathcal{L}_\varpi(\pi)} + \ \mbox{const}  \, .
\end{multline}
Policy optimization can then be understood as performing EM-style coordinate ascent on the function $\mathcal{F}(\pi, q_\varpi)$: First we find the distribution $\varpi^*$ that minimizes the KL divergence $D_{\mbox{\tiny KL}} \big( q_\varpi \ \vert\,\!\vert \ q_\pi \big)$, and then optimize the lower-bounding function $\mathcal{L}_{\varpi^*}(\pi)$ in \Cref{eq:Lvarpi}.
The lower-bounding function $\mathcal{L}_\mu(\pi)$ in \Cref{eq:Ma} is then obtained by assuming that $\mu$ is the policy obtained by the last iteration of the above updates (i.e., $\mu \approx \pi^{t-1}$), in which case $\varpi^* = \mu$ gives the optimal $q_{\varpi^*}$ above.

\subsection{Policy improvement objective} \label{app:policy_improv}
In \Cref{sec:penalty} we have shown that the multiplicative solution for the optimal policy $\pi$ can be re-written as:
\begin{align*}
\pi(a \mid x) &\propto \mu(a \mid x) \, \exp (r(x, a) / \beta) \, \\
&\propto \mu(a \mid x) \, \exp \big( (r(x, a) - g(x) ) / \beta \big)
\end{align*}
for any function $g(x)$. Here we present a derivation and show that when the function $g(x)$ is $\mathbb{E}_{a\sim \mu(\cdot \mid x)}[r(x, a)]$ which is the conditional expected reward under $\mu$ for context $x$, this corresponds to an objective function that is the lower bound on $J(\pi) - J(\mu)$, the gap on the expected reward between $\pi$ and $\mu$. 

Referring to the definition of $J(\pi)$ in \Cref{eq:policy_value}, note that for any function $g(x)$ we can equivalently optimize
\[
\max_\pi \mathbb{E}_{x\sim d(\cdot)}\mathbb{E}_{a\sim\pi(\cdot\mid x)}[r(x, a) - g(x)]
\]
When we set $g(x) = \mathbb{E}_{a\sim \mu(\cdot \mid x)}[r(x, a)]$ the above objective becomes
\begin{align}
 \E_{x \sim d(\cdot)} \, \E_{a \sim \pi(\cdot|x)} \, \big[ A_\mu(x,a) \big] = J(\pi) - J(\mu) \label{eq:policy_diff}
\end{align}
where $A_\mu(x,a)\defeq r(x, a) - \mathbb{E}_{a\sim \mu(\cdot \mid x)}[r(x, a)]$ as defined in \Cref{sec:penalty}. We can again introduce a KL divergence penalty between $\pi$ and $\mu$, averaged over the context $x\sim d(\cdot)$:
\begin{align}\label{eq:lowerbound_diff_direct}
J(\pi) - J(\mu) - \beta \cdot \mathbb{E}_{x\sim d(\cdot)}\Big[\kld{\pi(\cdot \mid x)}{ \mu(\cdot \mid x)}\Big] \leq J(\pi) - J(\mu)
\end{align}
That is the lower bound on $J(\pi) - J(\mu)$, the gap on the expected reward between $\pi$ and $\mu$. Plug \Cref{eq:policy_diff} into \Cref{eq:lowerbound_diff_direct}, we can re-write the lower bound (scaled by $1/\beta$) as:
\begin{align*}
&\mathbb{E}_{x\sim d(\cdot)}\mathbb{E}_{a\sim \pi(\cdot\mid x)}\Big[- \log \pi(a \mid x) + \big(\log \mu(a \mid x) + A_\mu(x, a)/\beta \big)\Big]\\
 =&~\mathbb{E}_{x\sim d(\cdot)}\Big[-\kld{\pi(\cdot \mid x)}{\mu(\cdot \mid x) \exp(A_\mu(x, a) / \beta)}\Big]
\end{align*}
In other words, maximizing the lower bound in \Cref{eq:lowerbound_diff_direct} is equivalent to minimizing the KL divergence $\kld{\pi(\cdot \mid x)}{\mu(\cdot \mid x) \exp(A_\mu(x, a) / \beta)}$ for all $x\sim d(\cdot)$. In an unconstrained setting, this leads to the following solution 
\[
\pi(a \mid x) \propto \mu(a \mid x) \, \exp (A_\mu(x, a) / \beta), \text{ for any } x\in \mathcal{X}.
\]
which is exactly what we want to derive. 

\subsection{Additional experimental results} \label{app:retailrocket_breakdown}
\begin{figure}
  \centering
    \begin{subfigure}{\columnwidth}
      \centering
  \includegraphics[width=\columnwidth]{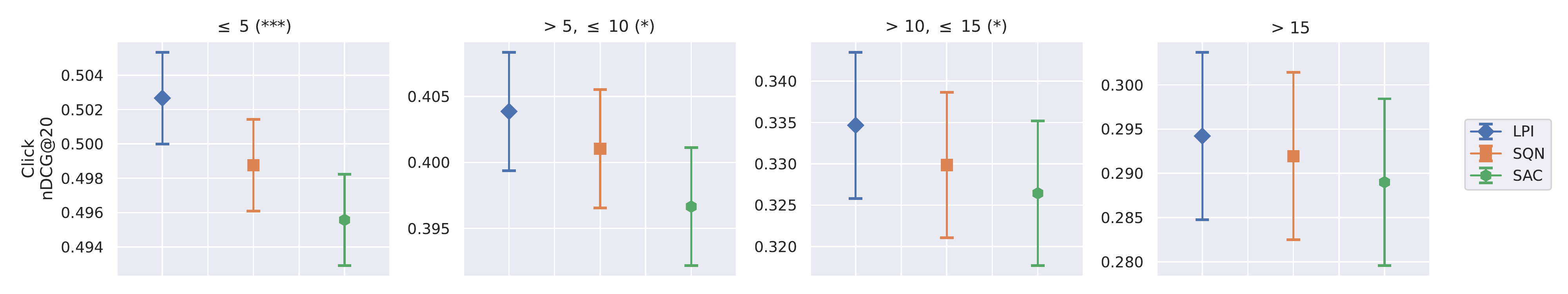}
        \caption{Sequence-level click nDCG by the number of clicks per sequence.}
        \label{fig:click_ndcg_retailrocket}
    \end{subfigure}
   \begin{subfigure}{\columnwidth}
     \centering
  \includegraphics[width=\columnwidth]{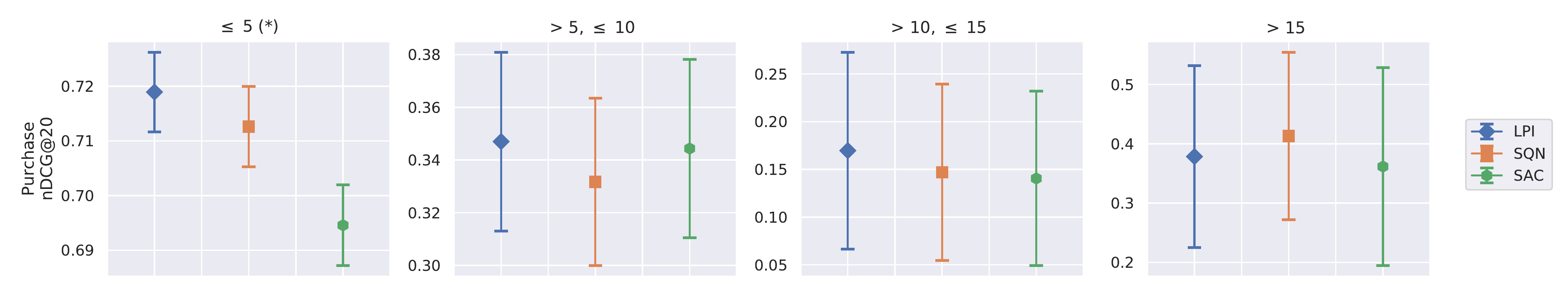}
        \caption{Sequence-lever purchase nDCG by the number of purchases per sequence.}
        \label{fig:purchase_ndcg_retailrocket}
    \end{subfigure}
    \caption{Sequence-level validation metric breakdown by action counts on RetailRocket. For each subplot, a paired t-test is performed and * indicates statistical significance at $\alpha=0.05$ level, ** at $\alpha=0.01$ level, and *** at $\alpha=0.001$ level.}
    \label{fig:ndcg_retailrocket_breakdown}
\end{figure}

In \Cref{fig:ndcg_retailrocket_breakdown} we show the validation metric breakdown on \textbf{RetailRocket}. Again we can see similar trend as observed in \Cref{fig:ndcg_yoochoose_breakdown}: The gains from \gls{LPI} are the most prominent on the bucket with $\leq$5 clicks/purchases, epecially with purchase.

